\documentclass[journal]{IEEEtran}
\usepackage{graphicx}
\usepackage{subcaption}
\usepackage{comment}
\usepackage{amsmath}
\usepackage{amssymb}
\usepackage{color}

\usepackage{verbatim}

\usepackage{fixltx2e}
\usepackage{hyperref}
\hypersetup{
  linkcolor=blue;
  citecolor=green;
}
\usepackage{color}

\begin{document}

\title{LGAN: Lung Segmentation in CT Scans  \\ Using Generative Adversarial Network}

\author{Jiaxing Tan*, Longlong Jing*, Yumei Huo, Yingli Tian$^\dag$, Fellow, \textit{IEEE}, Oguz Akin
 
\thanks{$^*$Equal contributions; $^\dag$Corresponding author.}

\thanks{Jiaxing Tan and Longlong Jing are with Department of Computer Science, the Graduate Center, City University of New York, New York, NY 10016, USA. }

\thanks{Yumei Huo is with Department of Computer Science, The College of Staten Island,  City University of New York, New York, NY 10016, USA. }

\thanks{Y. Tian is with the Department of Electrical Engineering, The City College, and the Department of Computer Science, The Graduate Center, The City University of New York, NY, 10031, USA. E-mail: ytian@ccny.cuny.edu.}
 
\thanks{Oguz Akin, MD is with Department of Radiology, Memorial  Sloan-Kettering (MSK) Cancer Center, New York, NY, 10065, USA. He receives support from MSK Cancer Center Support Grant/Core Grant (P30 CA008748).}
 }

\markboth{}%
{Shell \MakeLowercase{\textit{et al.}}: Bare Demo of IEEEtran.cls for IEEE Journals}

\maketitle

\begin{abstract}

Lung segmentation in computerized tomography (CT) images is an important procedure in various lung disease diagnosis. Most of the current lung segmentation approaches are performed through a series of procedures with manually empirical parameter adjustments in each step. Pursuing an automatic segmentation method with fewer steps, in this paper, we propose a novel deep learning Generative Adversarial Network (GAN) based lung segmentation schema, which we denote as LGAN. Our proposed schema can be generalized to different kinds of neural networks for lung segmentation in CT images and is evaluated on a dataset containing 220 individual CT scans with two metrics: segmentation quality and shape similarity. Also, we compared our work with current state of the art methods. The results obtained with this study demonstrate that the proposed LGAN schema can be used as a promising tool for automatic lung segmentation due to its simplified procedure as well as its good performance.

\end{abstract}


\maketitle

\section{Introduction}

\begin{figure*}[!t]
\centering
\includegraphics[width=6.5in]{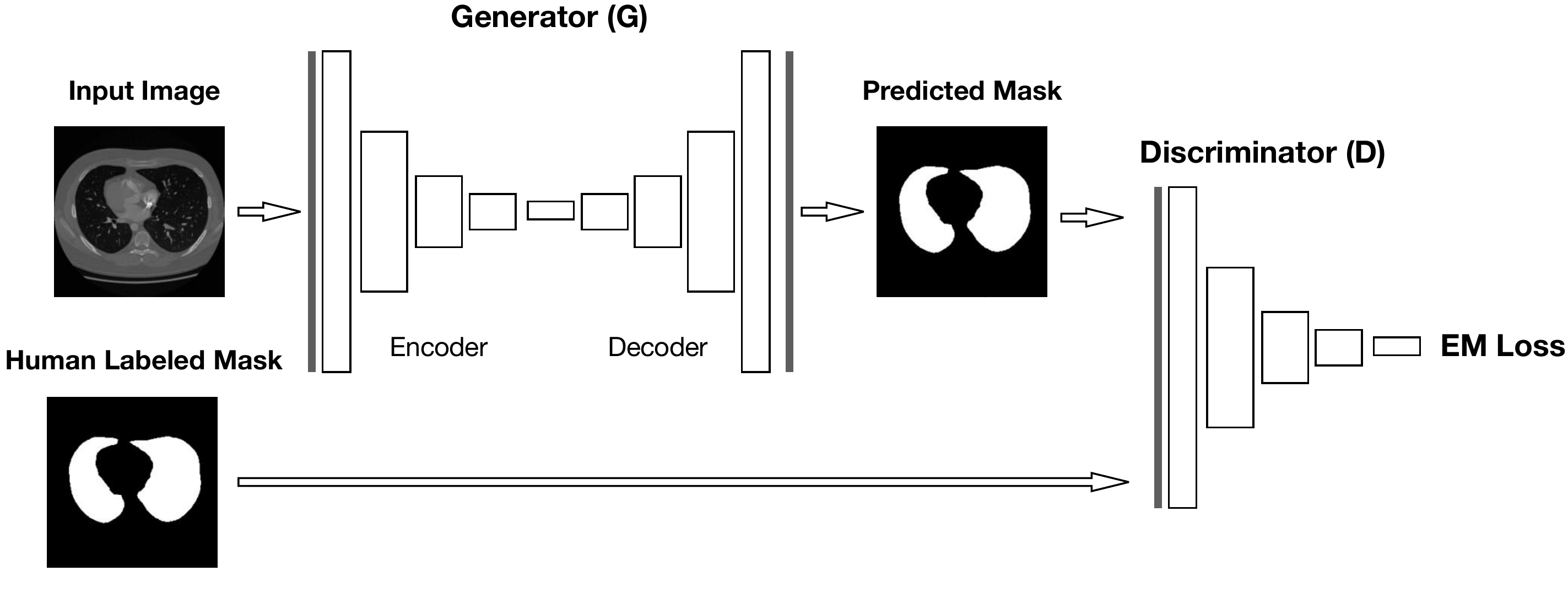}
\DeclareGraphicsExtensions.
\caption{The pipeline of the proposed LGAN schema which includes a generator network and a discriminator network. A Deep Deconvnet Network is trained to generate the lung mask while an Adversarial Network is trained to discriminate segmentation maps from the ground truth and the generator, which, in turn, helps the generator to learn an accurate and realistic lung segmentation of the input CT scans.}
\vspace{-15pt}
\label{fig_lgan}
\end{figure*}

Computerized tomography (CT) is a clinical imaging modality, that is key to, and sometimes the first step in, the diagnosis of various lung diseases, including lung cancer, which has been the leading cause of cancer-related deaths in the United States with the overall five-year survival rate of $17\%$ \cite{c1}. CT scan images allow a physician to confirm the presence of a tumor, measure its size, identify its precise location and determine the extent of its involvement with other nearby tissue. With an increasing use of CT imaging for diagnosis, treatment planning, and clinical studies, it has become almost compulsory to use computers to assist radiologists in clinical diagnosis and treatment planning.

Lung segmentation is an initial step in analyzing medical images obtained to assess lung disease. For example, early detection of lung nodules in lung cancer screening with CT is a tedious task for human readers and could be significantly improved with artificial intelligence. Researchers proposed a number of lung segmentation methods which fall into two categories: hand-crafted feature-based methods and deep learning-based methods. Compared to the hand-crafted feature-based methods, such as region growing \cite{adams1994seeded}, active contour model \cite{kass1988snakes}, and morphological based model \cite{mansoor2015segmentation}, deep learning-based methods (\cite{harrison2017progressive}, \cite{mansoor2015segmentation}, \cite{adams1994seeded}, \cite{kass1988snakes}, \cite{lalonde2018capsules}, \cite{zhao2018lung}) could automatically learn useful features \cite{lecun2015deep} without manually empirical parameter adjustments. 

Existing hand-crafted feature-based lung segmentation methods are usually performed through a series of procedures with manually empirical parameter adjustments. Various sets of 2D-based \cite{mansoor2015segmentation} and 3D-based methods \cite{sun2012automated} are developed to achieve a high quality segmented result.  

However, these traditional segmentation techniques are designed for specific applications, imaging modalities, and even datasets. They are difficult to be generalized for different types of CT images or various datasets since different kinds of features and different parameter/threshold values are extracted from different datasets. Moreover, the feature extraction procedure is monitored by users and the features/parameters are analyzed and manually and interactively adjusted.

In this paper, in contrast, we propose an end-to-end deep learning Generative Adversarial Network (GAN) based lung segmentation schema (LGAN), where the input is a slice of lung CT scan and the output is a pixel-wise mask showing the position of the lungs  by identifying whether each pixel belongs to lung or not. Furthermore, the proposed schema can be generalized to different kinds of networks with improved performance.

Recently, several deep learning-based pixel-wise classification methods have been proposed in computer vision area and some of them have been successfully applied in medical imaging. Early deep learning-based methods are based on bounding box \cite{Tan2017}. The task is to predict the class label of the central pixel(s) via a patch including its neighbors. Kallenberg \textit{et al.} \cite{kallenberg2016unsupervised} designed a bounding box based deep learning method to perform breast density segmentation and scoring of mammographic texture. Shin \textit{et al.} \cite{shin2016deep} compared several networks on the performance of computer-aided detection and proposed a transfer learning method by utilizing models trained in computer vision domain for medical imaging problem. Instead of running a pixel-wise classification with a bounding box, Long \textit{et al.} \cite{long2015fully}  proposed a fully convolutional network (FCN) for semantic segmentation by replacing the fully connected layers with convolutional layers. An Auto-Encoder alike structure has been used by Noh \textit{et al.} \cite{noh2015learning}  to improve the quality of the segmented objects. Later, Ronneberger \textit{et al.} \cite{ronneberger2015u} proposed a U-net model for segmentation, which consists of a contracting part as an encoder to analyze the whole image and an expanding part as a decoder to produce a full-resolution segmentation. The U-net architecture is different from \cite{long2015fully} in that, at each level of the decoder, a concatenation is performed with the correspondingly cropped feature maps from the encoder. This design has been widely used and proved to be successful in many medical imaging applications such as Lumbar Surgery\cite{baka2017ultrasound} and gland segmentation\cite{manivannan2017structure}. Most recently, Lalonde et al. \cite{lalonde2018capsules} designed a convolutional-deconvolutional capsule network, called SegCaps, to perform lung segmentation, where they proposed the concept of deconvolutional capsules. 

After the emergence of Generative Adversarial Network (GAN) \cite{GAN} based models, which have shown a better efficiency in leveraging the inconsistency of the generated image and ground truth in the task of image generation, Luc \textit{et al.} \cite{luc2016semantic} proposed a GAN based semantic segmentation model. The motivation is to apply GAN to detect and correct the high-order inconsistencies between ground truth segmentation maps and the generated results. The model trains a segmentation network along with an adversarial network that discriminates segmentation maps coming either from the ground truth or from the segmentation network. Following this idea, Zhao et al. \cite{zhao2018lung} proposed to use adversary loss to perform lung segmentation, where the segmentor is a fully convolutional neural network. 

Both of their works utilize the original GAN structure, which, however, due to its loss function design, original GAN suffers from the problem of learning instability such as {mode collapse, which means all or most of the generator outputs are identical}  (\cite{arjovsky2017wasserstein},\cite{arjovsky2017towards}).

To avoid this problem, Arjovsky et al. \cite{arjovsky2017wasserstein} proposed an optimized GAN structure which uses a new loss function based on the Earth Mover (EM) distance and in the literature is denoted as WGAN. It should be noted that WGAN is designed to solve the same problem as the original GAN, which is to leverage the inconsistency of the generated image and ground truth in the task of image reconstruction instead of generating an accurate segmentation from a given type of images.

In this paper, to solve the medical image segmentation problem, especially the problem of lung segmentation in CT scan images, we propose LGAN (Generative Adversarial Network based Lung Segmentation) schema which is a general deep learning model for segmentation of lungs from CT images based on a Generative Adversarial Network (GAN) structure combining the EM distance based loss function. In the proposed schema, a Deep Deconvnet Network is trained to generate the lung mask while an Adversarial Network is trained to discriminate segmentation maps from the ground truth and the generator, which, in turn, helps the generator to learn an accurate and realistic lung segmentation of the input CT scans. The performance analysis on the dataset from a public database founded by the Lung Image Database Consortium and Image Database Resource Initiative (LIDC-IDRI) shows the effectiveness and stability of this new approach. The main contributions of this paper include:

\begin{enumerate}
\item We propose a novel end-to-end deep learning Generative Adversarial Network (GAN) based lung segmentation schema (LGAN) with EM distance to perform pixel-wised semantic segmentation.

\item We apply the LGAN schema to five different GAN structures for lung segmentation and compare them with different metrics including segmentation quality, segmentation accuracy, and shape similarity. 

\item We perform experiments and evaluate our five LGAN segmentation algorithms as well as the baseline U-net model using LIDC-IDRI dataset with ground truth generated by our radiologists. 

\item Our experimental results show that the proposed LGAN schema outperforms current state-of-the-art methods on our dataset and debuts itself as a promising tool for automatic lung segmentation and other medical imaging segmentation tasks.

\end{enumerate}

\begin{figure}[!t]
\centering
\includegraphics[width=0.5\textwidth]{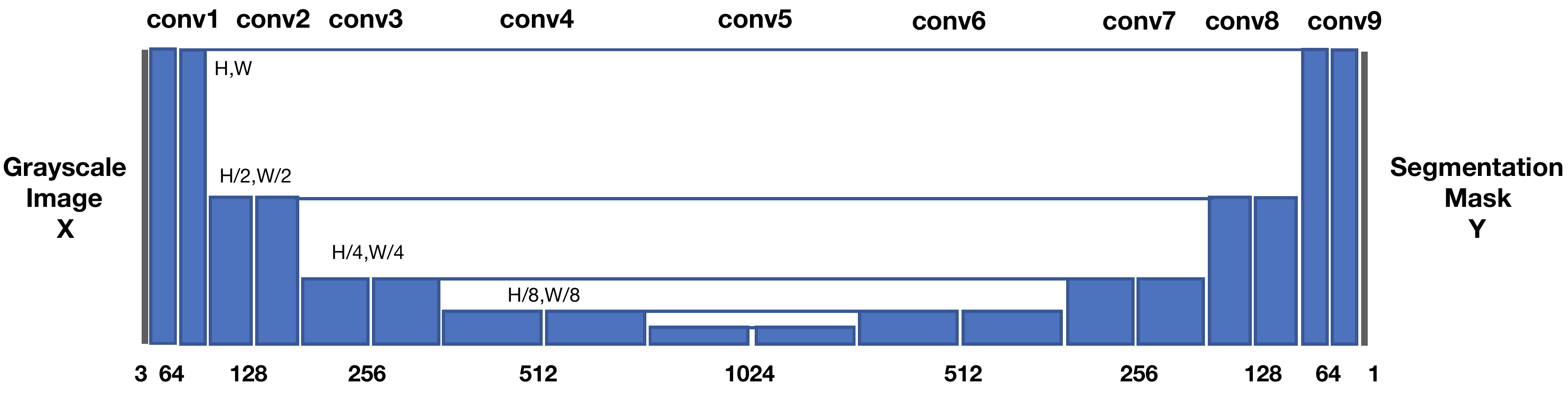}
 \DeclareGraphicsExtensions.
\caption{The architecture of the generator in the framework. Each blue box represents the feature map generated by convolution block. The number of channels is denoted on the bottom of the box. The lines on the top of the boxes indicate the concatenation operation of the feature map.}
\vspace{-10pt}
\label{generator}
\end{figure}

\section{The Proposed Method}

In this section, we first introduce the background knowledge of Generative Adversarial Network and then present the proposed LGAN schema.

\subsection{Generative Adversarial Networks}

Generative Adversarial Network (GAN) is a deep generative model initially proposed in \cite{GAN}, and later improved by DCGAN \cite{radford2015unsupervised} and WGAN \cite{arjovsky2017wasserstein}. A general GAN model consists of two kinds of networks named as the generator network and the discriminator network. The generator network is trained to generate an image similar to the ground-truth and meanwhile the discriminator network is trained to distinguish the generated image from the ground-truth image. By playing this two-player game, the results from the discriminator network help the generator network to generate more similar images and simultaneously the generated images as the input data help the discriminator network to improve its differentiation ability. Therefore, the generator network and the discriminator network are competing against each other while at the same time make each other stronger.

Mathematically, the goal of the generator network $G$ is to learn a distribution $p_z$ matching the ground-truth data in order to generate the similar data, while the goal of discriminator network $D$ is also to learn the distribution of the ground-truth data but for distinguishing the real data (i.e. from the real distribution $p_{data}$) from the generated data from $G$. The adversarial comes from the min-max game between $G$ and $D$, and is formulated as:
\begin{equation}
\small
\min_{G}\max_{D}V(D,G)=\mathbb{E}_{x \sim p_{data}(x)[log D(x)]}+ 
 \mathbb{E}_{z\sim p_{z}(z)[log(1-D(G(z)))]},
\end{equation}

where, for a given real data $x$ and the corresponding generated data $G(z)$, the adversarial discriminator is trained to maximize the probability output for the real data $x$ (that is, $\mathbb{E}_{x \sim p_{data}(x)[log D(x)]}$) and minimize the probability output for the generated data (that is, $\mathbb{E}_{x \sim p_{data}(x)[log D(G(z))]}$) which is equivalent to maximizing $\mathbb{E}_{z\sim p_{z}(z)[log(1-D(G(z)))]}$, and on the other side, the generator is trained to generate $G(z)$ as similar as possible to $x$ so that the discriminator outputs the bigger probability value for $G(z)$, that is, to maximize $\mathbb{E}_{x \sim p_{data}(x)[log D(G(z))]}$ and equivalently minimize $\mathbb{E}_{z\sim p_{z}(z)[log(1-D(G(z)))]}$.

Luc et al. \cite{luc2016semantic} employed GAN model to perform segmentation task, where the role of the generator has been changed from generating synthetic images to generating segmentation masks for the original images, which has been proved to be effective on the task of lung segmentation by Zhao et al \cite{zhao2018lung}. The details of GAN based segmentation design will be specified in the next section.

As illustrated in \cite{arjovsky2017wasserstein}, the original GAN structure, which although achieves a great performance in various tasks, including replicating images, human language, and image segmentation, suffers from a \text{mode collapse} problem due to its loss design. To make the training process more stable, Arjovsky \textit{et al.} proposed WGAN using Earth Mover (EM) distance to measure the divergence between the real distribution and the learned distribution \cite{arjovsky2017wasserstein}. Specifically, given the two distributions, $P_{data}$ and $P_z$, with samples $x \sim P_{data}$ and $y \sim P_z$, the EM distance is defined as: \begin{equation}
W(P_{data},P_z)=\inf_{\gamma \in \prod(P_{data},P_z)}\mathbb{E}_{(x,y)\sim  \gamma}[\left \| x-y \right \|],
\end{equation}

where $\prod (P_{data},P_z)$ represents the set of all joint distributions $\gamma(x, y)$ whose marginals are respectively $P_{data}$ and $P_z$, and the term $\gamma(x, y)$ represents the cost from $x$ to $y$ in order to transform the distributions $P_{data}$ into the distribution $P_z$. The EM loss actually indicates optimal transport cost. In this design, the loss for the generator network is:
\begin{equation}
    L_G = -\mathbb{E}_{x \sim P_z}[D(x)].
\end{equation}

And the loss for the discriminator network is:
\begin{equation}
    L_D = \mathbb{E}_{x \sim P_z}[D(G(x))]-\mathbb{E}_{x \sim P_{real}}[D(x)].
\end{equation}

\begin{figure}[!t]
\centering
\includegraphics[width=0.5\textwidth]{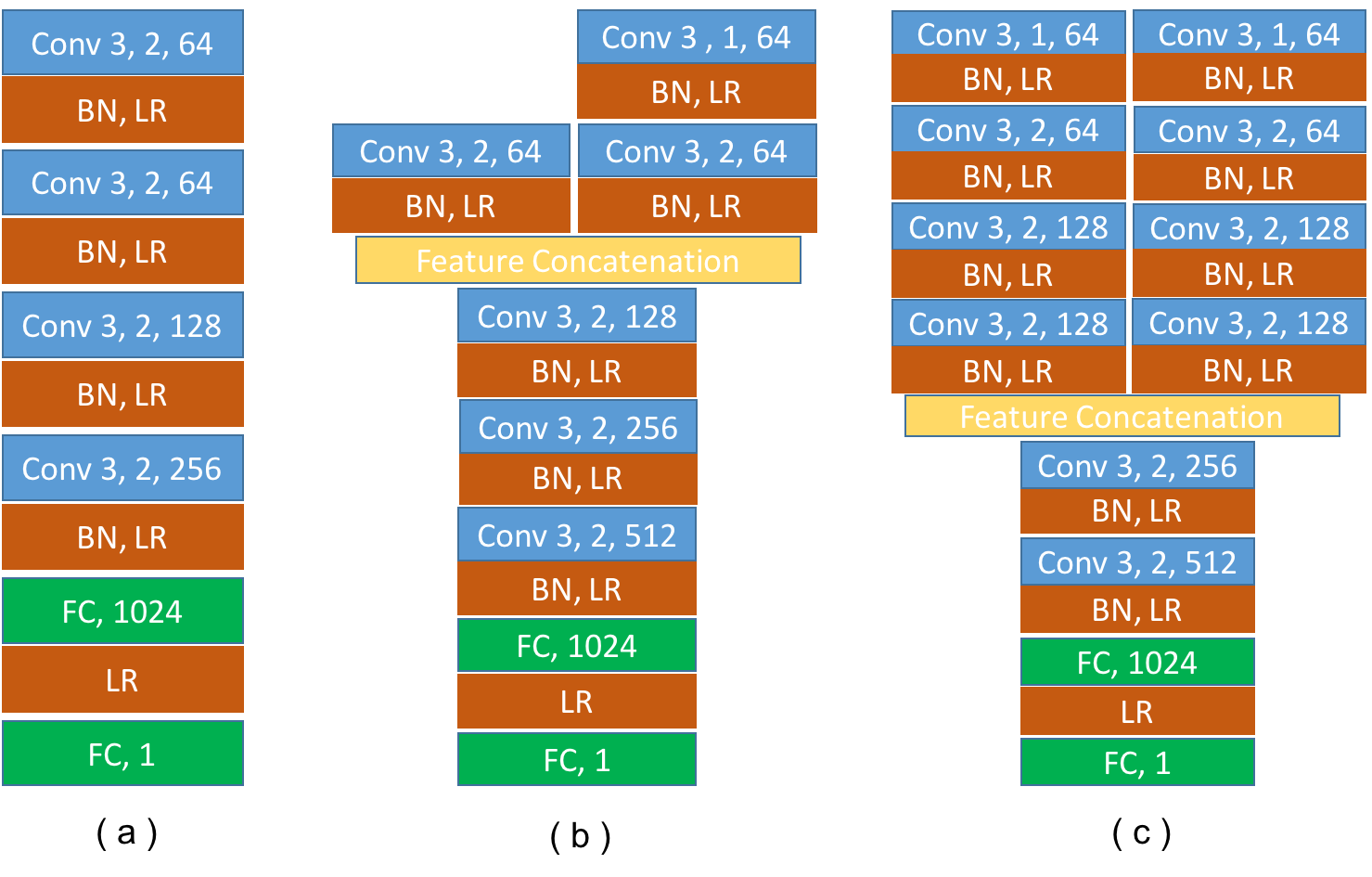}
\vspace{-10pt}
 \DeclareGraphicsExtensions.
\caption{The architecture of different $D$ networks, where \textit{Conv} stands for convolution layer, \textit{FC} stands for fully-connected layer, \textit{BN} stands for Batch Normalization and \textit{LR} stands for LeakyReLU. For each convolution layer, the numbers represent kernel size, (down) pooling stride and number of kernels accordingly. Feature concatenation layer combines feature maps from different branches.}
\vspace{-10pt}
\label{fig:Discriminator}
\end{figure}

With the EM distance based loss, the GAN model becomes more powerful in generating high-quality realistic images and outperforms other generative models. While the WGAN is designed for image reconstruction, here we take advantage of the basic idea of WGAN, and design an efficient and enhanced deep learning Generative Adversarial Network based Lung Segmentation (LGAN) schema.

\subsection{Our LGAN Schema}
Our Generative Adversarial Network based Lung Segmentation (LGAN) schema is designed to force the generated lung segmentation mask to be more consistent and close to the ground truth and its architecture is illustrated in Fig. \ref{fig_lgan}. 

LGAN consists of two networks: the generator network and the discriminator network, and both of them are convolution neural networks. The generator network is to predict the lung segmentation masks based on the grayscale input images of CT slices, while the discriminator computes the EM distance between the predicted masks and the ground truth masks.

During the training and testing, the LGAN schema takes a slice of the lung CT scan $I_i$ as input, then the generator predicts a mask $M_i$ to illustrate the pixels belong to the lung. The quality of lung segmentation is judged by how well $M_i$ fools the discriminator network. In the rest of this section, we describe the three main components of our LGAN schema: Generator Network, Discriminator Network, and Training Loss.

\subsubsection{Generator Network}

The generator network is designed to generate the segmented mask of the input lung CT scan image. The mask labels all the pixels belonging to the lung. This segmentation task can be addressed as a pixel-wise classification problem to identify whether a pixel belongs to the lung area or not. Given an input CT slice $I_i$, the generator will predict the category of each pixel and generate a corresponding mask $M_i$ based on the classification result.

The architecture of our designed generator is illustrated in Fig.~\ref{generator}. The generator model consists of encoder and decoder parts. The encoder extracts the high-level features from the input CT scan by a bunch of convolution blocks, while the decoder reconstructs the mask from the high-level features. Both encoder and decoder are composed of convolution blocks, which are represented as blue boxes in the figure. The input of the generator network is normalized to $224 \times 224$ pixels and the generated mask is the same size as the input.

In the encoder part, each block has two convolution layers, both of which have the same number of filters with filter size $3 \times 3$ followed by a max-pooling layer,  which performs a $2\times2$ down-pooling on the feature map.

In the decoder part, each block consists of one deconvolution layer and two convolution layers. For the convolution layers, similarly, each has the same number of filters with filter size $3 \times 3$. Instead of an up-pooling layer, we use the deconvolution layer with stride $2$ as suggested by Tran \textit{et al.} \cite{tran2016deep} because deconvolution layer can generate better quality images than an up-pooling layer. {Following other work, we add skip connection between the encoder and decoder.}

Following DCGAN \cite{radford2015unsupervised}, we employ LeakRelu as the activation function for the convolution layers which is first proposed in \cite{maas2013rectifier}. As shown below, to alleviate potential problems caused by ReLU, which sets $0$ to all negative values, LeakyReLU set a small non-zero gradient NegativeSlope, which is user pre-defined, to negative values. In the equation below, we represent this negative slope as $\alpha$.
\begin{equation}
\label{eq.evenly}
LeakyReLU(x,\alpha)= max(x,0)+ \alpha \times min(x,0).
\end{equation}
At the final layer, a $1*1$ convolution is performed to map each component feature vector to the final segmentation mask.

\begin{table*}[t]%
 \centering
\caption{The list of all the proposed five LGAN structures and their corresponding descriptions.}
\label{tab:Models}
\begin{center}
\begin{tabular}{|c|c|c|c|}
  \hline
  Network                  & Input \\
  \hline
  {$LGAN_{Basic}$}  &  generated mask, one at a time.\\
  \hline
  { $LGAN_{Product}$}&segmented original image based on the predicted mask.\\
  \hline
   $LGAN_{EF}$ &mask and original image are combined as one input with two channels.   \\ 
  \hline
  $LGAN_{LF}$ & mask and original image as two inputs.\\
  \hline

 $LGAN_{Regression}$    & generated mask and ground truth, so as to approximate EM loss directly.   \\
  \hline
\end{tabular}
\end{center}
\centering
\vspace{-10pt}
\end{table*}%

\subsubsection{Discriminative Network}\label{DNet}

The task of the discriminative network is to distinguish the ground truth mask from the generated segmentation mask. We employ the EM distance to measure the difference between the real and the learned distributions as it has been proved to be a smooth metric \cite{arjovsky2017wasserstein}. 

Given a generator,  the discriminator approximates $E[X]$ function such that the EM loss $E[X]-E[Y]$ is approximated by $D(G(z))-D(Real)$. Compared to the discriminator in the original GAN, which performs a classification task, the new discriminator is actually performing a regression task (approximating the function $E(X)$). 

Based on the different assumptions that could help improve the performance of the discriminator network, we propose five different designs for the discriminator network, which thus yield five different LGAN structures. The details of the five frameworks are demonstrated in Table \ref{tab:Models}. We discuss these five designs one by one in the next section.

\subsubsection{Training Loss}\label{GLoss}

As the original WGAN is designed for image reconstruction tasks, here we modify the training loss to fit for the segmentation task. Specifically, we modify the loss of generator $G$ by adding a Binary Cross Entropy (BCE) loss which calculates the cross-entropy between the generated mask and ground truth mask. Therefore, the loss of the generator network is:
\begin{equation}
    \mathbb{BCE}[G(x),Real]-\mathbb{E}_{x \sim P_z}[D(x)],
\end{equation}
Where $P_z$ is the learned distribution from the ground-truth mask by $G$.

For the training loss of the discriminator $D$, different designs for the discriminator network may have the different training loss functions which are described in the next section.

\section{The Proposed LGAN Structures}\label{FiveLGAN}

In GAN-based image reconstruction tasks, the generated image and the ground truth image are very similar. However, for lung segmentation task, the pixel intensity in the generated mask is in $[0, 1]$ while the value in the ground truth mask is binary, that is, either $0$ or $1$. This fact may mislead the discriminator to distinguish the generated mask and the ground truth mask by simply detecting if the mask consists of only zeros and ones (one-hot coding of ground truth), or the values between zero and one (output of segmentation network).

With this observation, we explore all possible discriminator designs for lung segmentation task based on various assumptions, and provide five different LGAN structures:  LGAN with Basic Network, LGAN with Product Network, LGAN with Early Fusion Network, LGAN with Late Fusion Network and LGAN with Regression Network. Their corresponding architectures are illustrated in Fig.~\ref{fig:Discriminator}. In the rest of this section, we introduce them accordingly.

\subsection{LGAN with Basic Discriminator}

The basic discriminator is to evaluate the generated mask and the ground truth mask separately and minimize the distance between the two distributions, which can be illustrated as (a) in Fig.~\ref{fig:Discriminator}. We denote the LGAN with this basic discriminator as $LGAN_{Basic}$ and it has a single channel with the network input size of $224\times224$.

In $LGAN_{Basic}$, the training loss of $G$ is the same as we described in section \ref{GLoss} and the training loss of $D$ is the following:
\begin{equation}
    \mathbb{E}_{x \sim P_z}[D(G(x))]-\mathbb{E}_{x \sim P_{real}}[D(x)].
\end{equation}
Based on $LGAN_{Basic}$, we conjecture that the discriminator network may have a more precise evaluation if the original image is also provided as additional information. Under this assumption, we examine three strategies and design the $LGAN_{Product}$, $LGAN_{EF}$, and $LGAN_{LF}$ structures.

\begin{table}[!t]
\renewcommand{\arraystretch}{1.3}
\caption{ Performance comparison of different LGAN structures.}
\label{table_example}
\centering
\begin{tabular}{|c|c|c|c|c|}
 \hline
 &\multicolumn{2}{|c|}{Mean} & \multicolumn{2}{|c|}{Median} \\
\hline
Model &IOU  & Hausdorff &IOU  & Hausdorff\\
\hline
$Benchmark$ &0.6248  & 6.1062 & 0.7582  & 5.8310\\
\hline
$LGAN_{Basic}$ &0.9018  & 3.3672 & 0.9655  & 3.1622\\
\hline
 $LGAN_{Product}$ &0.8858  & 3.4095 & 0.9635 & 3.3166\\
\hline
$LGAN_{EF}$ &0.9015  & 3.2836 & 0.9656  & 3.1622\\
\hline
$LGAN_{LF}$ &0.7911  & 3.5318 & 0.9444  & \textbf{3.0}\\
\hline
$LGAN_{Regression}$ &\textbf{0.9225}  & \textbf{3.3802} & \textbf{0.9715}  & 3.1622\\
\hline
\end{tabular}
\end{table}

\subsection{LGAN with Product Network}

Different from the basic discriminator where the inputs are the segmented mask and the ground-truth mask with only binary value, the product network takes, as the inputs, the lung volume images which are mapped out from the original CT scan image by the segmented mask and the ground-truth mask respectively. That is, given the segmentation mask, we obtain the lung volume image by modifying the original image such that the values within the segmented lung area are kept as they are but the values in the rest of the area are set to be 0. This design is motivated by the work of Luc et al. \cite{luc2016semantic}.

With this input, the discriminator network might be biased by the value distribution. Although in \cite{luc2016semantic} the deep learning model with the product network is not designed based on WGAN, we observe that the product network could still be used in our LGAN model and we define this LGAN structure as $LGAN_{Product}$. Note that the discriminator in $LGAN_{Product}$ differs from $LGAN_{Origin}$ only in inputs, so the discriminator in $LGAN_{Product}$ shares the same architecture as $LGAN_{Origin}$ shown in (a) of Fig.~\ref{fig:Discriminator}. 

In $LGAN_{Product}$, the loss of G is the same as we described in section \ref{GLoss} and the training loss of D, which is product network, is the following:
 
\begin{equation}
    \mathbb{E}_{x \sim P_z}[D(G(x)\circ I_i)]-\mathbb{E}_{x \sim P_{real}}[D(x)],
\end{equation}
 
where $\mathit{x} \circ \mathit{y}$ is a pixel-wise multiplication of matrix $\mathit{x}$ and $\mathit{y}$, and the mask and the original image has the same size. 

\subsection{LGAN with Early Fusion Network}

Instead of taking only the mask as inputs, early fusion network takes both the whole original CT scan image and the segmentation/ground-truth mask as an input. To keep the design of single input, we concatenate the original image and the mask as one single image with two channels, where one channel is the original CT scan and the other is the mask. We denote $LGAN$ with this early fusion discriminator network as $LGAN_{EF}$. 

The architecture of the discriminator in $LGAN_{EF}$ is shown in (a) of Fig.\ref{fig:Discriminator}. Different from $LGAN_{Basic}$ and $LGAN_{Product}$,  $LGAN_{EF}$ has the input size of $224\times224$ with 2 channels which are the concatenation of the original CT scan and its mask. In $LGAN_{EF}$, the training loss of G is the same as we described in section \ref{GLoss} and the loss of the discriminator network is:
\begin{equation}
    \mathbb{E}_{x \sim P_z}[D(G(x)\bigoplus I_i)]-\mathbb{E}_{x \sim P_{real}}[D(x)\bigoplus I_i],
\end{equation}

where $\mathit{x} \bigoplus \mathit{y}$ represents the operation of concatenating matrix $\mathit{x}$ and $\mathit{y}$ into a single matrix with 2 channels.

\subsection{LGAN with Late Fusion Network}

Another way of taking both the original image and the mask as an input in the discriminator network is to employ the late fusion technique. Specifically, the input of the discriminator is the concatenation of the high-level feature of the CT scan and the mask. We denote LGAN with this type of discriminator as $LGAN_{LF}$. 

The corresponding architecture of the discriminator in $LGAN_{LF}$ is shown in (b) of Figure \ref{fig:Discriminator}. There are two branches of convolution operations in this discriminator. One is for the CT scans, and the other is for the masks. The two inputs first pass the two branches separately, and then their features are fused by a concatenate layer and pass through several convolution layers and down-sampling layers before they pass through the fully-connected layers to reach the final result. As the CT scan is more complicated than the mask to be understood by the network, we let the CT scan pass through more convolution layers before it features are concatenated. 

In $LGAN_{LF}$,  the training loss of $G$ is the same as we described in section \ref{GLoss} and the loss of the discriminator network is:
\begin{equation}
    \mathbb{E}_{x \sim P_z}[D(G(x),I_i )]-\mathbb{E}_{x \sim P_{real}}[D(x, I_i)].
\end{equation}

\subsection{LGAN with Regression Network}

As we addressed in section~\ref{DNet}, in WGAN, the EM loss $E[X]-E[Y]$ is approximated by $D(G(z))-D(Real)$ where $D(G(z))$ and $D(Real)$ are evaluated separately and independently. Differently, we design the discriminator as a regression network to approximate the $E[D(G(z))]-E[D(Real)]$ where $D(G(z))$ and $D(Real)$ are evaluated together in the same network setting. The regression discriminator network takes two inputs, the real mask and the mask generated by the generator network. The output of the network is the approximated EM distance, and the network is optimized by minimizing the distance. We denote LGAN with this regression discriminator network as $LGAN_{Regression}$. 

The architecture of the discriminator in $LGAN_{Regression}$ is shown in (c) of Figure \ref{fig:Discriminator}. Similar to the previous networks, the inputs in the regression discriminator network first separately pass through their own convolution branches and down-sampling layers before their features are concatenated together. And then the concatenated features pass through more convolution layers before getting into the fully-connected layer. In this discriminator network, the convolution branch consists of a long set of individual convolution layers.

In $LGAN_{Regression}$, the loss of $G$ is the same as we described in section \ref{GLoss} and the loss of the discriminator network is:
\begin{equation}
    \mathbb{E}_{x_1 \sim P_z,x_2 \sim P_{real}}[D(G(x_1),x_2 )].
\end{equation}

{By playing the min-max game, the generator prevents the distance computed by discriminator from going to positive infinity while the discriminator network prevents it from going to negative infinity. The generator and the discriminator networks play this min-max game for several rounds until a tie is reached.}

\section{Experiments}

\subsection{Dataset}

The original RAW CT Scan images are acquired from the public database founded by the Lung Image Database Consortium and Image Database Resource Initiative (LIDC-IDRI), as we have previously used in \cite{han2015fast}. 

There are 220 patients' CT Scans in our dataset, and each scan contains more than 130 slices. Each CT slice has a size of $512 \times 512$ pixels. We randomly select 180 patients' scans as the training data and the other 40 patients' scans as the testing data for experiments. To generate the ground-truth segmentation mask for lung segmentation, on each CT scan, we first apply the Vector Quantization based Lung segmentation method as described in \cite{han2015fast} to filter out major lung parts with region growing applied to smooth the result. Then the mask is further corrected by radiologists.

\subsection{Experiment Design}

Our proposed methods of five different LGAN structures are validated and compared on LIDC dataset. The comparison on structures are described in section~\ref{FiveLGAN} and are listed in TABLE~\ref{tab:Models}.

Furthermore, our best model is compared with the state-of-the-arts for lung segmentation task on LIDC-IDRI dataset following the same settings and evaluation metrics.

At last, as our method could serve as a pre-processing step for nodule detection. Therefore, three cases with lung nodules located close to the lung boundary are investigated to understand if our generated masks could include those nodules.

We use Adam \cite{kingma2014adam} optimizer with a batch size of $32$. All models are trained from scratch without using pretrained weights. The learning rate is set to $10^{-5}$, momentum to $0.9$, and weight decay to $0.0005$. The network is initialized with a Gaussian Distribution. During testing, only the Segmentor network is employed to generate the final mask. The source code will be made publicly available on the project website following the acceptance of the paper.

\subsection{Evaluation Metrics}

We take two metrics to evaluate the performance of the networks: segmentation quality and shape similarity.

\subsubsection{Segmentation Quality}

Intersection over Union (IOU) score is a commonly used for semantic segmentation. Given two images $X$ and $Y$, where $X$ is the predicted mask and $Y$ is the ground truth. The IOU score is calculated as:

\begin{equation}
    IOU=\frac{X\cap Y}{X\cup Y},
\end{equation}

which is the proportion of the overlapped area to the combined area. 

\subsubsection{Shape Similarity}

To evaluate the similarity between shapes, the commonly used Hausdorff distance \cite{rockafellar2009variational} is employed to measure the similarity between the segmented lung and the ground truth. In this paper, we use the symmetrical Hausdorff distance mentioned in \cite{nutanong2011incremental} as the shape similarity evaluation metric.

Given generated mask $ \mathfrak{M}$ and groundtruth $ \mathfrak{G}$, the symmetrical Hausdorff distance is calculated as:
\begin{equation}
\small
HausDist(\mathfrak{M},\mathfrak{G})=\max\begin{Bmatrix}
  \sup_{x\in \mathfrak{M} } \inf_{y\in \mathfrak{G}} \left \| x-y \right \| ,
\\ 
 \sup_{x\in \mathfrak{G} } \inf_{y\in \mathfrak{M}} \left \| x-y \right \|  
\end{Bmatrix}.
\end{equation}

For all three evaluation metrics, we compute and compare their mean values as well as their median values.

\section{RESULTS}

\subsection{Comparison results of our proposed different structures}

\begin{figure}[!t]
\centering
\includegraphics[width=0.5\textwidth]{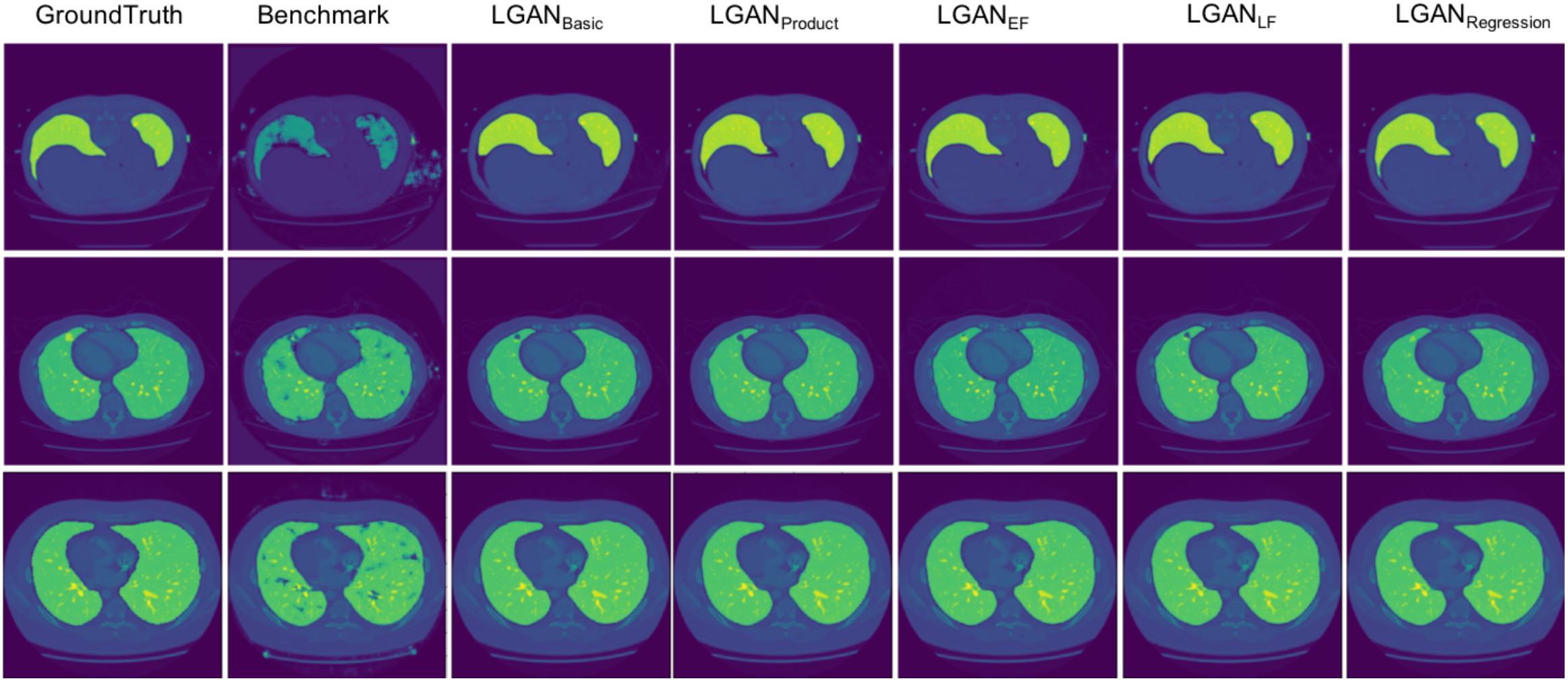}
\vspace{-10pt}
\DeclareGraphicsExtensions.
\caption{Segmentation results of different LGAN structures on 3 lung CT slices. The first column is the ground truth masks of the lung areas on the CT slices. The green color represents the lung areas, while the blue color represents the other areas. The other columns are the segmentation results of different LGAN structures. }
\vspace{-15pt}
\label{fig_comp}
\end{figure}

The experimental results of LGAN are shown in TABLE~\ref{table_example}. The LGAN models achieve a significant improvement in the performance compared with U-Net, which is a segmentation network that has the same architecture as the generator network. The performance of LGAN is more than $20$\% higher than the benchmark, which demonstrates the effectiveness of the LGAN. All the LGAN achieve better performance than the generator alone, among which the $LGAN_{Regression}$ obtains the best performance.

To qualitatively study the performance of the proposed architectures and demonstrate the strength of our proposed GAN framework, we compare the performance of all the models on three CT slices. As shown in Fig.~\ref{fig_comp}, the significant improvement in lung segmentation using LGAN structures can be observed. All of the LGAN models can capture the lung area, while the $LGAN_{Regression}$ achieves the best result.

\begin{figure}[!t]
\centering
\includegraphics[width=0.5\textwidth]{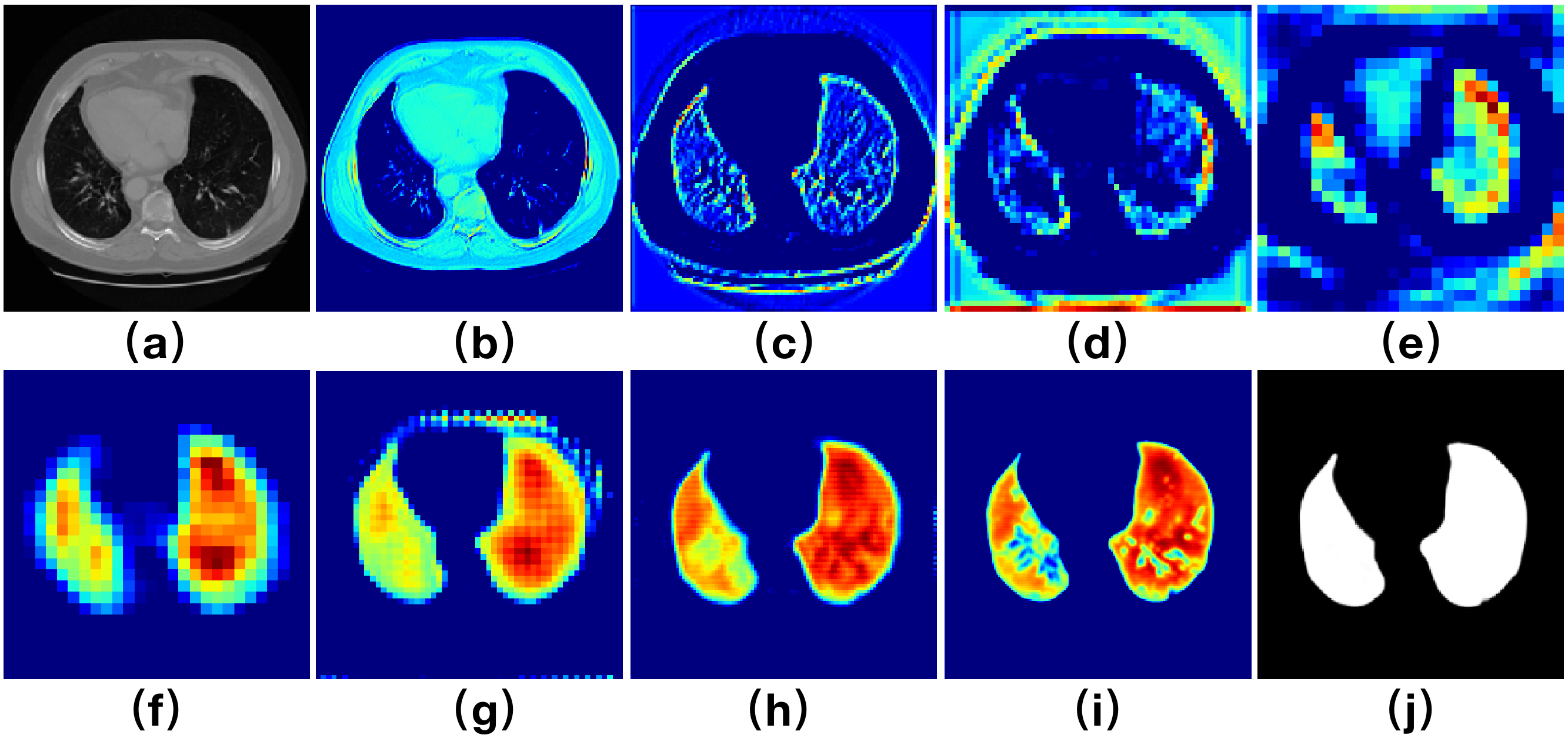}
 \DeclareGraphicsExtensions.
 \vspace{-5pt}
\caption{Visualization of activations of the generator network. The activation maps from (b) to (i) correspond to the output maps from lower to higher layers in the generator. We select the most representative activation in each layer for effective visualization. The image (a) is the input image and image (k) is the predicted mask. The finer details of the lung are revealed, as the features are forward-propagated through the layers in the generator. It shows that the learned filters tend to capture the boundary of the lung.}
\vspace{-5pt}
\label{fig_fmap}
\end{figure}

One problem of the neural network based methods is that it is hard to know what really happened inside the network. Therefore, besides segmentation accuracy, we would like to attain further insight into the learned convolution models. We select the most representative feature maps obtained by each layer of the generator. The feature maps of LGANs are illustrated in Fig.~\ref{fig_fmap}. The feature maps show that the network can extract the major information about lung boundary through the contracting part, and then gradually expanding the extracted highly compressed features into a clear mask. Obviously, the lung area tends to have a brighter color, which means higher activation, than the rest parts of the image.

\subsection{Comparison with State-of-the-Arts}

We compare the performance of our LGAN\textsubscript{Regression} model with the current state-of-the-arts of lung segmentation on LIDC-IDRI dataset, including the traditional benchmark method \cite{mansoor2015segmentation}, U-net model \cite{ronneberger2015u}, Tiramisu Network\cite{jegou2017one}, and SegCaps \cite{lalonde2018capsules}. The commonly used 3D Dice-score metrics and the mean as well as median values are calculated following the same settings.  As shown in TABLE ~\ref{workcomp}, our model achieves the highest score comparing to current state-of-the-arts with an average Dice-score of 0.985 and a median Dice-score of 0.9864. Although SegCaps \cite{lalonde2018capsules} claims to have fewer parameters, the designed capsule is very memory consuming. Also, our model is much shallower than the 100-layer Tiramisu model \cite{jegou2017one} and achieves better performance. Meanwhile, our model outperforms the gNet \cite{zhao2018lung}, which utilizes the original GAN loss. Comparing to the traditional methods, such as Morph, which requires a series of thresholding, morphological operations, and component analysis, our end-to-end model provides an one-step solution.  Moreover, as mentioned in \cite{zhao2018lung}, due to the high-resolution of the LIDC-IDRI CT volumes, 0.001 better Dice-score indicates that 5k more pixels are correctly covered.

\begin{table}[!t]
\renewcommand{\arraystretch}{1.3}
\caption{Performance Comparison with the state-of-the- arts (3D Dice-score).}
\label{workcomp}
\centering
\begin{tabular}{|c|c|c|}
 \hline
 Model &Mean& Median\\
\hline

Morph \cite{mansoor2015segmentation} &0.862$\pm$2.93 & - \\
\hline
U-net \cite{ronneberger2015u} &0.970$\pm$0.59 & 0.98449 \\
\hline
Fusion gNet \cite{zhao2018lung} &0.983$\pm$0.05 & - \\

\hline
Tiramisu \cite{jegou2017one} & -  & 0.9841 \\
\hline
SegCaps \cite{lalonde2018capsules} & - & 0.9847 \\
\hline
LGAN\textsubscript{Regression} & \textbf{0.985$\pm$0.03}  & \textbf{0.9864}\\
\hline
\end{tabular}
\end{table}

\subsection{Case Study}
\begin{figure}[!t]
\centering
\includegraphics[width=0.5\textwidth]{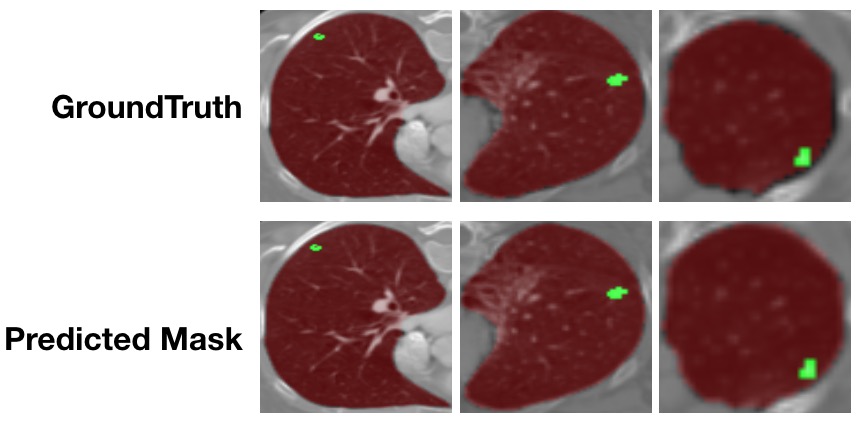}
 \DeclareGraphicsExtensions.
\caption{Examples demonstrate our models  are able to segment nodules which are close to lung boundary. The areas in Brown are the segmented lung and the green areas are the lung nodules.}
\vspace{-15pt}
\label{fig_nodule}
\end{figure}

As lung segmentation usually serves as a pre-processing step for many tasks such as lung nodule detection, we investigate whether the segmented lung areas by our model include all nodules even when the nodules are very close to lung boundary. As shown in fig.~\ref{fig_nodule}, our method can include all the nodules inside the lung area besides achieving high-quality lung mask. 

\subsection{Discussion}
For our LGAN schema, we have designed five different discriminative networks, and evaluate these five structures with experiments on lung segmentation. The experimental results demonstrate that our proposed LGAN structures significantly outperform the basic GAN framework, which shows the effectiveness of the LGAN schema. Our model outperforms current state-of-the-arts of segmentation task on LIDC-IDRI dataset with higher Dice-score. Furthermore, our method can be an effective step for lung nodule detection task and can be applied to the nodules on lung boundary.

The generator network in our LGAN model is designed based on the currently most widely used benchmark method, U-Net. As the task of finding an optimal network structure is still ongoing, our LGAN schema could also be optimized correspondingly. A deeper network design would extract higher-level features but requires more data as well as more parameters and higher computation cost. Patch normalization and random initialization in our model training show a significant effect and the optimization method by Adam \cite{kingma2014adam} is also used in our work.

\section{CONCLUSIONS}
Lung segmentation is usually performed by methods such as thresholding and region growing. Such methods, on the one hand, require dataset-specific parameters, and on the other hand, require a series of pre- and post-processing to improve the segmentation quality. To reduce the processing steps for lung segmentation and eliminate the empirical based parameters adjustments, we have proposed a Generative Adversarial Network based lung segmentation schema (LGAN) by redesign the discriminator with EM loss. The lung segmentation is achieved by the adversarial between the segmentation mask generator network and the discriminator network which can differentiate the real mask from the generated mask. Such adversarial makes the generated mask more realistic and accurate than a single network for image segmentation. Moreover, our schema can be applied to different kinds of segmentation networks.

\bibliographystyle{IEEEtran}
\bibliography{bare_jrnl}
  
\end{document}